\documentclass{article}

\usepackage{arxiv}

\usepackage[utf8]{inputenc} % allow utf-8 input
\usepackage[T1]{fontenc}    % use 8-bit T1 fonts
\usepackage{hyperref}       % hyperlinks
\usepackage{url}            % simple URL typesetting
\usepackage{booktabs}       % professional-quality tables
\usepackage{amsfonts}       % blackboard math symbols
\usepackage{nicefrac}       % compact symbols for 1/2, etc.
\usepackage{microtype}      % microtypography
\usepackage{lipsum}		% Can be removed after putting your text content
\usepackage{graphicx}
\usepackage{doi}

\usepackage{hyperref}
\hypersetup{
    colorlinks=true,
    linkcolor=blue,
    filecolor=magenta,      
    urlcolor=cyan,
}
\usepackage{url}            % simple URL typesetting
\usepackage{booktabs}       % professional-quality tables
\usepackage{nicefrac}       % compact symbols for 1/2, etc.
\usepackage{microtype}      % microtypography
\usepackage{xcolor}
\usepackage{multirow}
\usepackage{graphicx}

\usepackage{amsmath,amsfonts,bm,amssymb,amsthm}
\usepackage{enumitem}
\usepackage{arydshln}
\usepackage[linesnumbered,ruled,lined,boxed]{algorithm2e}
\SetKwInput{KwInput}{Input}
\SetKwInput{KwOutput}{Output}
\usepackage{footnote}
\makesavenoteenv{algorithm}

\def\N{\mathbb{N}}
\def\R{\mathbb{R}}

\def\O{\mathcal{O}}

\title{Efficient GPU implementation of randomized SVD and its applications}

\date{September 1, 2021}

\author{ \L{}ukasz Struski \\
	Faculty of Mathematics and Computer Science \\ 
	Jagiellonian University, Krak\'ow, Poland \\
	\texttt{lukasz.struski@uj.edu.pl} \\
	\And
	Pawe\l{} Morkisz \\
	NVIDIA Corp., Santa Clara, USA \\[0.5em]
	AGH University of Science and Technology,\\
	Faculty of Applied Mathematics, Krak\'ow, Poland \\
	\texttt{pmorkisz@nvidia.com} \\
	\And
	Przemys\l{}aw Spurek \\
	Faculty of Mathematics and Computer Science \\ 
	Jagiellonian University, Krak\'ow, Poland \\
	\texttt{przemyslaw.spurek@uj.edu.pl} \\
	\And
	Samuel Rodriguez Bernabeu \\
	NVIDIA Corp., Santa Clara, USA \\
	\texttt{srodriguezbe@nvidia.com} \\
	\And
	Tomasz Trzci\'nski \\
	Faculty of Mathematics and Computer Science \\ 
	Jagiellonian University, Krak\'ow, Poland \\
	\texttt{tomasz.trzcinski@uj.edu.pl} \\
}

\begin{document}
\maketitle

\begin{abstract}

Matrix decompositions are ubiquitous in machine learning, including applications in dimensionality reduction, data compression and deep learning algorithms. Typical solutions for matrix decompositions have polynomial complexity which significantly increases their computational cost and time. In this work, we leverage efficient processing operations that can be run in parallel on modern Graphical Processing Units (GPUs), predominant computing architecture used e.g. in deep learning, to reduce the computational burden of computing matrix decompositions. More specifically, we reformulate the randomized decomposition problem to incorporate fast matrix multiplication operations (BLAS-3) as building blocks. We show that this formulation, combined with fast random number generators, allows to fully exploit the potential of parallel processing implemented in GPUs.
Our extensive evaluation confirms the superiority of this approach over the competing methods and we release the results of this research as a part of the official CUDA implementation\footnote{https://docs.nvidia.com/cuda/cusolver/index.html}.

\end{abstract}

% \linenumbers

\section{Introduction}

Matrix decomposition is a fundamental operation widely used across numerous real-life applications, including data compression, optimization of machine learning models and dimensionality reduction. The latter application remains especially challenging, given the increasing amount of real-world highly-dimensional data, such as multi-lingual textual corpora or drug discovery databases with complex chemical compounds. 
%Real-world data often has a large number of attributes (high-dimensional structures) for instance, texts in natural language processing are embedded in high-dimensional space, in order to preserve information about their semantics. Similarly in chemistry, where compounds are represented by their structural features. 
Although the original dimensionality of those datasets is high, in practice the data lays on lower-dimensional subspace with a smaller number of parameters. Dimensionaly reduction methods, such as singular value decomposition (SVD)~\cite{wall2003singular}, are typically used to find this lower-dimensional subspace in real-life problems ranging from
% In practice, such data lay on a smaller dimensional subspace, with a much smaller number of parameters than the number of features. 
% For decades, dimensional reduction problems have been identified in many applications and domains, such as 
data analysis~\cite{ding2010application,harkat2006improved}, data compression~\cite{dony2001karhunen,gastpar2006distributed}, statistics and machine learning ~\cite{li2000recursive,hoffmann2007kernel} as well as clustering~\cite{STRUSKI2018161}. % where the task is to look for a lower-dimensional linear subspace that minimizes the sum of squared distances to the points. 
Despite the ubiquity of SVD, its computational complexity poses a significant problem, especially when dealing with large-scale datasets.

% method for dimensionality reduction. Although every year we increase computational power, we struggle with efficiently processing large-scale datasets using traditional matrix algorithms. 
One of the approaches proposed to address this limitation of SVD is to approximate matrices of low rank using randomization techniques~\cite{feng2018faster,oropeza2010randomized}.
These techniques are well known to give a significant speed-up, yet their practical implementations are often cumbersome and do not leverage recent advancements in the architectural design of contemporary computational hardware, such as Graphical Processing Units (GPUs). For instance, the baseline methods used to solve this task, such as GESVD or Krylov subspace methods~\cite{krylov1931cislennom}, do not utilize computational linear algebra to parallelize the operations, and effectively reduce the computational cost of SVD. %The SVD algorithm has three steps: the reduction of the general dense matrix to bidiagonal form, the extraction of the singular values from the condensed form and the accumulation of the orthogonal transformations to form the singular vectors. The bidiagonalization step is often the bottleneck of the  SVD algorithm because it requires multiple matrix multiplications. Krylov methods are bounded by the performance of matrix-vector multiplication (GEMV) and hence bounded by the speed at which we can access the data. In the case of GESVDR, its performance depends on the problem size, the portion of the spectrum, and the number of the power method iterations. Some combination of these parameters will make it behave like a GEMM, a tall and skinny QR, or a standard singular value decomposition.

In this work, we present a new approach towards solving randomized SVD problem efficiently by exploiting fast matrix multiplications and random number generators, implemented on contemporary GPUs. More specifically, we rely on the computational linear algebra to decompose randomized SVD into a set of low-level matrix multiplications known as Basic Linear Algebra Subprograms\footnote{BLAS functionality is categorized into three sets of routines called "levels": BLAS-1 operations typically take linear time $\O(n)$, Bias-2 operations quadratic time, and BLAS-3 operations cubic time.} (BLAS)~\cite{lawson1979basic}. These subprograms contain BLAS-1 and BLAS-2 operations, which are limited by system bandwidth, and BLAS-3 operations bounded by the arithmetic operation throughput of a machine. 
% To study the performance of algorithms we typically model these BLAS-2 and BLAS-3 operations.
The key to achieving high performance matrices operations is the effective translation of the computations into a set of BLAS-3 operations, since those operations can be easily run in parallel on modern hardware architectures such as GPUs. %Throughput-oriented processors, like GPUs, are especially well suited for this kind of operation.
This paper shows that such a solution can be efficiently implemented on the contermporary GPUs and allows for faster processing of large-scale dataset. % Graphical Processing Units (GPUs). Thanks to this approach we are able to process larger data set in a reasonable time, still obtaining the prescribed accuracy

% Baseline methods that are used to solve this task are GESVD or Krylov subspace methods~\cite{krylov1931cislennom}, to understand why the randomized SVD can outperform those, we need to understand two factors: the asymptotic complexity of an algorithm, and how efficiently a given algorithm can be mapped to a given hardware architecture.

% The algorithmic complexity is a measure of the number of arithmetic operations (FLOPs, floating-point operations) a given algorithm performs for a given input size. The algorithmic complexity is the factor that dominates the run time of the algorithm for large problem sizes, however, for small or medium problems, is it sometimes more important how efficiently we can perform these floating-point operations. The first factor does not depend on the platform on which we execute our algorithm, whereas the latter greatly depends on the hardware. For example, a matrix-vector multiplication has a complexity of $\O(n^2)$, but the speed at which we can execute these operations is limited by how fast we can read the entries of A, hence it is bounded by the bandwidth of our memory subsystem.

\begin{figure}[!ht]
    \centering
    \includegraphics[width=\textwidth]{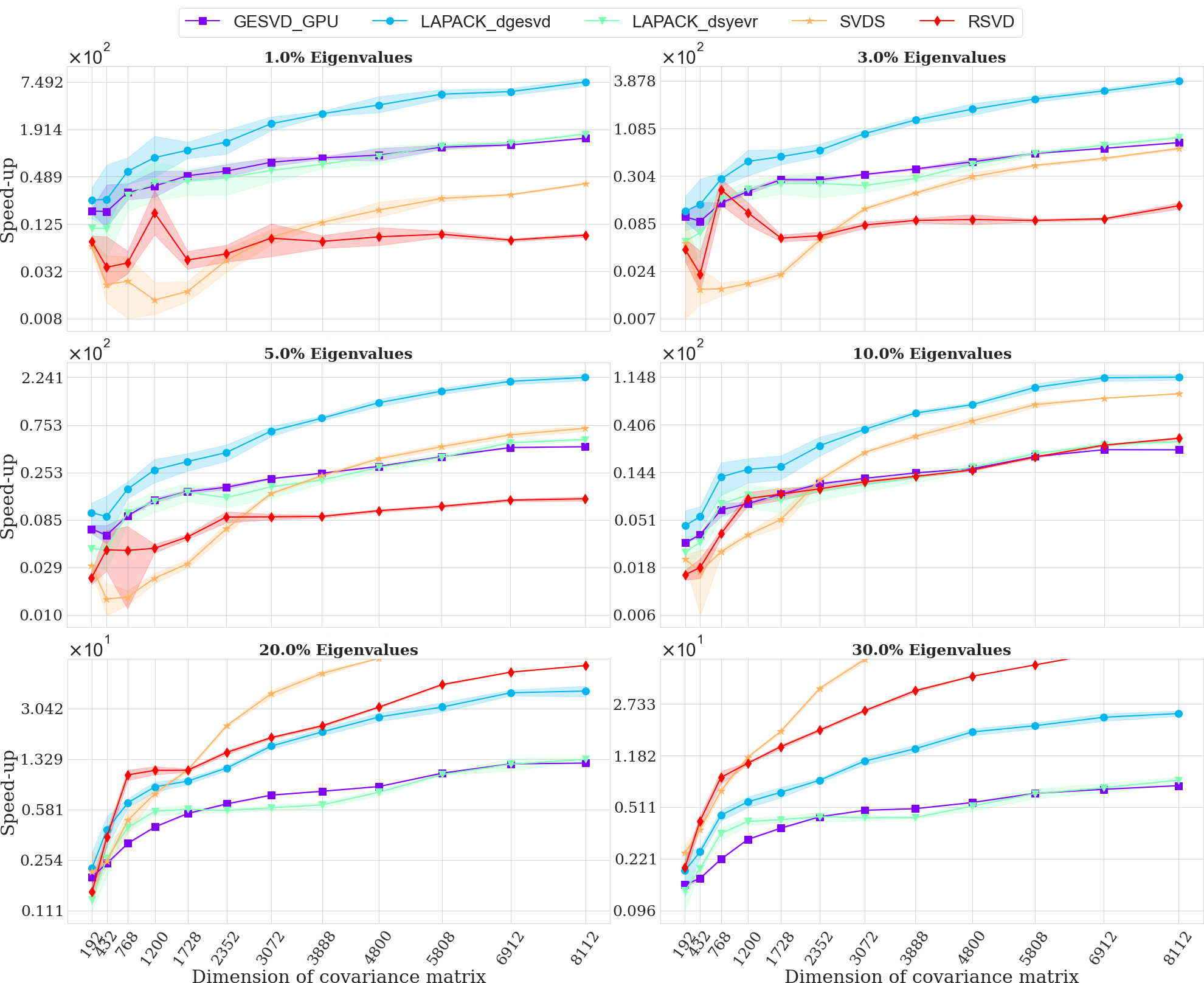}
    \caption{Average time and standard deviation of 10 attempts to run competing methods in relation to run time of our method (speed-up). We consider two types of methods that calculate the whole spectrum and that calculate only $k$ largest eigenvalues.}
    \label{fig:speedup_pca}
\end{figure}

% This phenomenon is exploited in computational linear algebra in particular. Basic Linear Algebra Subprograms\footnote{BLAS functionality is categorized into three sets of routines called "levels": BLAS-1 operations typically take linear time $\O(n)$, Bias-2 operations quadratic time, and BLAS-3 operations cubic time.} (BLAS)~\cite{lawson1979basic} consists of BLAS-1 and BLAS-2 operations, which are limited by the bandwidth of our system, and BLAS-3 operations, which are bounded by the throughput at which our machine can perform arithmetic operations. To study the performance of algorithms we typically model these BLAS-2 and BLAS-3 operations.
% The key to achieving high performance matrces operations is to effectivnes use the computations into BLAS-3 operations because these operations are suitable for the parallel hardware architecture of GPUs. Throughput-oriented processors, like GPUs, are especially well suited for this kind of operation.

To confirm the efficacy of our method, we extensively evaluate our approach against the state-of-the-art competitors. %This paper shows that a probabilistic matrix algorithm can be effectively applied to Graphical Processing Units. 
Figure~\ref{fig:speedup_pca} shows the results of this evaluation which confirms the effectiveness of our implementation and a significant acceleration it obtains relative to state-of-the-art methods. In the presented experiment, we consider two types of methods that calculate the whole spectrum of eigenvalues such as GESVD GPU and LAPACK~\cite{anderson1999lapack} dgesvd, as well as the methods that calculate only $k$ largest eigenvalues such as LAPACK dsyevr, randomized singular value decomposition (RSVD)~\cite{rsvd} and SVDS from package \cite{RSpectra}. As a dataset, we take CelebA~\cite{liu2015faceattributes} where we resize the images to different sizes and our approach shows consistent 10x speedup, cf. Section~\ref{sec:experiments}. 

To summarize, our contributions can be defined as follows:
\begin{itemize}
    \item We introduce a novel approach for speeding up randomized SVD computations based on fast GPU operations: BLAS-2 matrix multiplications and random number generations. %based on efficient  approach provides good accuracy for top-k eigenvalues.
    \item We extensively evaluate our method against state-of-the-art approaches and empirically prove its improved efficacy. %In practice works out of the box for k no greater than 10\% of the data dimension.
    \item We make our findings available to the public by releasing our code as part of the official CUDA implementation.
\end{itemize}

The remainder of this work is structured in the following manner. Section~\ref{sec:related} discusses related works. In Section~\ref{sec:method} we introduce our method and in Section~\ref{sec:experiments} we evaluate it against competitors. We conclude this work in Section~\ref{sec:conclusions}.
% \todo{related work / State of the art - regarding eigenvalues estimation (e.g. Lanczos algorithm)}

\section{Related works}
\label{sec:related}

Principal component analysis (PCA) is commonly used for dimensionality reduction by projecting each data point onto only the first few principal components to obtain lower-dimensional data while preserving as much of the data's variation as possible. 
It can be shown that the principal components are eigenvectors of the data's covariance matrix. Thus, the principal components are often computed by eigendecomposition of the data covariance matrix or singular value decomposition (SVD) of the data matrix.

If the matrix is small, we can compute eigenvalues using the characteristic polynomial. In practice, eigenvalues of large matrices are not computed using the characteristic polynomial, and algorithms to find them are iterative. There are Iterative numerical algorithms for approximating roots of polynomials, such as Newton's method, but, in general, it is impractical to compute the characteristic polynomial and then apply these methods. One reason is that small round-off errors in the coefficients of the characteristic polynomial can lead to large errors in the eigenvalues and eigenvectors~\cite{trefethen1997numerical}.

One of the simplest and most accurate iterative algorithms is the power method \cite{mises1929praktische} (also known as the Von Mises iteration).  In this method, a given vector is repeatedly applied to a matrix and properly normalized. Consequently, it will lie in the direction of the eigenvector associated with the eigenvalues. The power iteration is simple, but it may converge slowly. Moreover,  the method approximates only one eigenvalue of a matrix. The Power approach is usually used as a part of more efficient techniques, e.g. Krylov methods \cite{krylov1931cislennom}, inverse iteration, QR-method. The basis of the QR method for calculating the eigenvalues of $A$ is the fact that $n \times n$ real matrix can be factorized as $A=QR$ where $Q$ is orthogonal and $R$ is upper triangular. The method is efficient for the calculation of all eigenvalues of a matrix.

In the case of a large data set, we do not usually require all the eigenvalues, but only a few first ones. Partial singular value decomposition algorithms are usually based on the Lanczos algorithm \cite{calvetti1994implicitly,larsen1998lanczos,lehoucq1998arpack,baglama2005augmented}. Randomized matrix algorithms nowadays are most popular in practical applications. In \cite{frieze2004fast} authors proposed Monte Carlo approach to SVD decomposition. The method for approximation of low-rank SVD decomposition is based on non-uniform row and column sampling. 
In \cite{sarlos2006improved,liberty2007randomized,martinsson2011randomized}  authors introduced a more robust approach based on random projections. The method constructs a subspace that captures the column space of a matrix. 
The modification of the method, which uses fast matrix multiplications was introduced in \cite{woolfe2008fast}. In \cite{halko2011finding} authors unified and expanded previous work on the randomized singular value decomposition and introduced state-of-the-art algorithms to compute the approximation of low-rank singular value decomposition.

The basic idea of probabilistic matrix algorithms is to apply a degree of randomness to derive a smaller matrix from a high-dimensional matrix, which captures the essential information. Then, a deterministic matrix factorization algorithm is applied to the smaller matrix to compute a near-optimal low-rank approximation. 
Recent has been demonstrated that this probabilistic framework can be effectively used to compute the pivoted QR decomposition~\cite{duersch2017randomized}, the pivoted LU decomposition~\cite{shabat2018randomized}.

\section{Method description}
\label{sec:method}

In this section, we define the problem and introduce our method that leverages GPU computations of probabilistic matrix operations to speed up the processing time. 

First, let $A\in\R^{m\times n}$ be an arbitrary matrix for any $m, n\in\N$, and $k\in\N$ be the target number of singular values. Recall that making singular value decomposition (SVD) of $A$ we obtain
\[
A = U\cdot\Sigma\cdot V^T,
\]
where $U\in\R^{m\times r}$, $V\in\R^{n\times r}$ are orthogonal matrices for $r=rank(A)\leq \min(m, n)$. 
The above decomposition of $A$ is called as the {\em compact SVD} while it is a simple transformation of a classical SVD decomposition
\[
A = \bar{U}\cdot\bar{\Sigma}\cdot \bar{V}^T = 
\begin{bmatrix}
U & \hat{U}
\end{bmatrix} 
\cdot
\begin{bmatrix}
\Sigma & 0 \\
0 & 0
\end{bmatrix} 
\cdot
\begin{bmatrix}
V^T \\
\hat{V}^T
\end{bmatrix} 
= U\cdot\Sigma\cdot V^T,
\]
where $\bar{U}\in\R^{m\times m}$, $\bar{V}\in\R^{n\times n}$ are unitary and $\hat{U}\in\R^{m\times m - r}$, $\hat{V}\in\R^{n\times n - r}$ have orthonormal columns.

% where $U=\left[u_1, \cdots, u_r\right]\in\R^{m\times r}$, $V=\left[v_1, \cdots, v_r\right]\in\R^{n\times r}$ are orthogonal matrices for $r=rank(A)\leq \min(m, n)$.
% 
% For $k\leq r$ we define
% \[
% A_k := \sum_{i=1}^k\sigma_{ii}u_iv_i^T,
% \]
% where $\sigma_{11}\geq\ldots\geq\sigma_{rr}>0$ are the singular values. 
% We are only interested in the top $k\ll r$.

In our approach, we use the randomized SVD algorithm as it is a straightforward realization of the proposed randomized SVD code from \cite{halko2011finding}, which computes the $k$-SVD of $A$ up to $1 + \varepsilon$ Frobenius norm relative error. 
Let $C = A\cdot S\in\R^{m \times s}$ be a good sketch of $A$, the column space of $C$ should roughly contain the columns of $A_k$ by the low-rank approximation property i.e.
\[
A_k := \underset{B}{argmin}\|A - B\|^2_F \quad\text{subject to}\quad rank(B)\leq k,
\]
where $\|.\|_F$ is Frobenius norm and $s\geq k$.
If $S$ is a Gaussian projection matrix \cite{johnson1984extensions} or count sketch \cite{charikar2002finding} and {$s = \lceil\O(\frac{k}{\varepsilon})\rceil$}\footnote{The notation $\O$ defines the asymptotic behavior.}, then the low-rank approximation property
\[
    \min_{rank(B)\leq k}\|CB - A\|^2_F\leq (1 + \varepsilon) \|A - A_k\|^2_F
\]
holds in expectation.

From the above property we can describe randomized $k$-SVD pseudo-algorithm in the several steps that are described in Algorithm~\ref{alg:kSVD}.

\begin{algorithm}[h]\label{alg:kSVD}
	\SetAlgoLined
	\KwInput{$A\in\R^{m\times n}$, a target number $k$ of singular vectors, an
			exponent $q$ (usually, $q\in\{1, 2\}$), and a small number $p\in\N$.}
	
	For $s = k + p$ draw a sketching matrix $S\in\R^{n\times s}$\tcp*[r]{the time complexity $\O(ns)$ ($S$ is just a Gaussian matrix)}
	
 	Compute $Y=(A\cdot A^*)^q\cdot A\cdot S$ for a given $q$\tcp*[r]{the time complexity $\O(mnsq)$}
	
	% Compute $q$ steps of QR iteration $Y=(A\cdot A^*)^q \cdot A \cdot S$\tcp*[r]{the time complexity \change{$\O(mnsq)$}}
	
	Construct a matrix $Q$ whose columns form an orthonormal basis for the range of $Y$ by computing the QR decomposition of $Y = Q\cdot R$ using QR decomposition\footnote{The QR decomposition of a matrix $A\in\R^{m\times n}$ into a product $A = Q\cdot R$ of an orthogonal matrix $Q\in\R^{m\times n}$ and an upper triangular matrix $R\in\R^{n\times n}$.}\tcp*[r]{the time complexity is $\O(msk)$}
	
	$B=Q^T\cdot A\in\R^{s\times n}$\tcp*[r]{the time complexity $\O(mns)$}
	
	$B = U\cdot\Sigma\cdot V^T$\tcp*[r]{decomposition SVD, the time complexity $\O(ns^2)$}
	
	$\tilde{U} = Q\cdot U\in\R^{m\times k}$\tcp*[r]{the time complexity is $\O(msk)$}
	
	\KwOutput{$A_k\approx\tilde{U}\cdot\Sigma\cdot V^T$}
	\caption{Randomized $k$-SVD pseudo-algorithm to solve partial eigenproblem of matrix $A$ ($k$-th largest eigenvalues of matrix $A$).}
\end{algorithm}

% ---- description 

The description of the proposed algorithm can be found below.

Step 1: Using a custom number generator, we draw the sketching matrix directly on the GPU. This process is parallel and compute-bound.

Steps 2 and 3: These steps consist of a QR factorization of a tall and skinny factorization (GEQRF) where the matrix $Q$ is explicitly formed (ORGQR) and applied to the input matrix $A$ employing matrix-matrix multiplication (GEMM).
The tall-and-skinny QR is bandwidth bound and benefits from the higher bandwidth of the GPU. ORGQR forms the matrix Q explicitly using a BLAS-3 rich formulation \cite{schreiber1987}. Applying the orthogonal matrix to the input matrix $A$ is a BLAS-3 operation, highly efficient on the GPU.

Step 4: This step can be seen as a different version of matrix-matrix multiplication (GEMM).

Step 5: This step consists of a singular value decomposition of a small matrix. The size of this matrix is significantly smaller than the input matrix (order of the target rank+ some oversampling). Our implementation uses the Jacobi eigenvalue solver in cusolver \cite{cusolver_doc}. This solver can offer more parallelism than the QR algorithm. At its core, the performance of the QR algorithm is dominated by the diagonalization step (SYTRD), which usually accounts for up to $60$--$70\%$ of the total runtime, and it is rich in BLAS-2 operations. On the other hand, the Jacobi solver can offer more parallelism by applying up to $n/2$ non-overlapping Givens rotations in parallel. A small number of iterations, usually $3$--$4$, are required because it converges quadratically.

Traditionally, the computation of small matrices has been dominated by CPUs. The cost of invoking GPU kernels and moving data from main memory to device memory (and back) made GPU computations less attractive. However, a combination of faster PCI data transfers and strategies to limit the kernel launch overhead have moved this threshold to smaller and smaller matrices ({\it e.g.}, for a $500\times500$ matrix: MKL: $0.02514$s and CUDA GESVDJ (Jacobi-based SVD): $0.00021795$s measured on NVIDIA TESLA V100 and Intel(R) Core(TM) i9-10920X CPU @ 3.50GHz with MKL 2020.0.166). In this particular case, the data transfer is amortized over the various steps of the algorithm. In terms of kernel launch overhead, the latency-bound routines in this algorithm have been highly optimized over the last few years up to the point where some of them only require a single kernel launch.

Step 6: Finally, one more GEMM computes the left eigenvectors if required.

Note that $\Sigma\in\R^{k\times k}$ so if you want only $k$-th largest eigenvalues of matrix $A$, we used only the first five points of the Algorithm~\ref{alg:kSVD} procedure, i.e. we needed only the matrix $\Sigma$. In Section~\ref{sec:experiments} we describe and show results of experiments, where we calculate only $k$-th largest eigenvalues.

%\todo{Is this necessary?

%Power iteration. 
%\begin{enumerate}
%    \item Input: $m \times n$ matrix $A$, target rank $k$, and desired precision $\epsilon$
%\end{enumerate}

%This procedure approximates the rank-$2k$ factorization $U\Sigma V^*$, where $U$ and $V$ are orthonormal, and $\Sigma$ is nonnegative and diagonal.

%----- TODO: add some information about the computational complexity of each step and the summarized computation complexity

%Complexity of method:

%\begin{itemize}
%    \item Sketch: $\omega = A \times S$. Performs $\O(mn)$
%    \item $C = A \times \Omega$. Performs $\O(mns)$
%    \item $Q = QR(C)$. Performs $\O(ms^2)$
%    \item GESVD. Performs $\O(k^3)$
%    \item Backtransform; Performs $\O(mnk)$
%\end{itemize}
%}

\paragraph{GPU computations to speed up processing} The above formulation let us decompose the problem of SVD into a set of matrix operations (predominantly multiplications) with a significant number of random numbers induced by the probabilistic approach. Exactly those two features of our formulation lead to a significant speed up observed when implementing Algorithm~\ref{alg:kSVD} on modern GPU architectures.

Thanks to the parallel nature of GPU processing, matrix-matrix operations can be effectively computed using BLAS-3 operations.
% as much as possible as those by nature of matrix operations are suitable for the parallel hardware architecture of GPUs. 
Throughput-oriented processors, such as GPUs, are especially well suited for this kind of operation. BLAS-3 kernels break down linear algebra operations in terms of GEMM-like operations, which attain very high performance on graphic processors.

Moreover, random number generators effectively optimized for GPU architecture can offer up to threefold speedup in processing time thanks to the CuRAND library\footnote{https://developer.nvidia.com/curand}. Combined with the decomposition of randomized SVD into a set of matrix-matrix multiplications our approach leads to a significant computation speedups, as we present in the following sections.

\section{Experiments}\label{sec:experiments}

In this section, we evaluate the performance of our approach and compare it to other methods available either in CPU or in GPU. We document three different experiments: in the first one, we consider matrix A that is not square, then we test application of randomized SVD to principal component analysis of a random matrix, and finally we use our method to practical method of subspace clustering~\cite{STRUSKI2018161}.

In the first two experiments, we run every method 10 times on the same datasets and calculate for average time denoted by $\mathrm{mean(\cdot)}$ and standard deviation denotes by $\mathrm{std(\cdot)}$. In the Figures we use the solid lines to draw the ratio $\frac{\mathrm{mean(*)}}{\mathrm{mean(\textit{our\_method})}}$, the dashed black line has a value of 1 (a reference to our method), and the shaded region of error defined by interval:
\[
\left[
\tfrac{\mathrm{mean(*)} - \mathrm{std(*)}}{\mathrm{mean(\textit{our\_method})} + \mathrm{std(\textit{our\_method})}}; \tfrac{\mathrm{mean(*)} + \mathrm{std(*)}}{\mathrm{mean(\textit{our\_method})} - \mathrm{std(\textit{our\_method})}}
\right],
\]
where $*$ denotes another method.

We consider two types of methods that calculate the whole spectrum such as GESVD GPU and LAPACK~\cite{anderson1999lapack} dgesvd and methods that calculate only $k$ largest eigenvalues such as LAPACK dsyevr, randomized singular value decomposition (RSVD)~\cite{rsvd} and SVDS from package \cite{RSpectra}.

\begin{figure}[!ht]
    \centering
    \includegraphics[width=\textwidth]{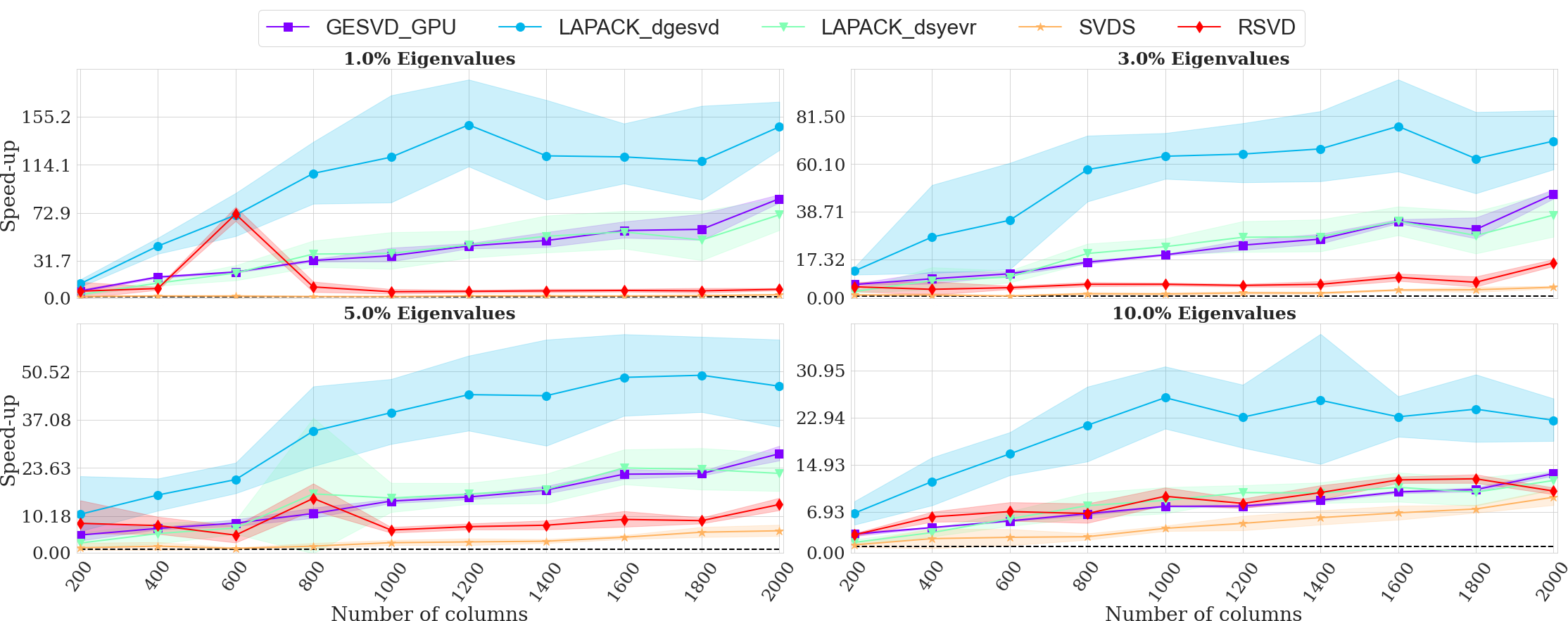}
    \caption{The speed-up of other methods to ours in the 'fast decay' case of comparison. We show mean values of 10 runs as the line with standard deviation as the light areas. In this case we create matrix $A^{2000\times n}$, where $n$ is the number of columns and we calculate 1\%, 3\%, 5\%, 10\% of the largest eigenvalues for it.}
    \label{fig:perf_comp_decay_fast}
\end{figure}

\paragraph{Performance comparison} In this experiment, we make three different test cases for which we construct the matrix $A = U\cdot\Sigma\cdot V^T\in\R^{m\times n}$ for $m\geq n > 0$ with random orthogonal matrices $U\in\R^{m\times m}, V\in\R^{n\times n}$ and a certain spectrum $\Sigma = [\sigma_{ij}]\in\R^{m\times n}$, where $\sigma_{ij} = 0$ for $i\not = j$ and
\begin{enumerate}[label=(\roman*)]
    \item $\sigma_{ii} = \frac{1}{i^2}$ --- fast decay,
    \item $\sigma_{ii} = 0.0001 + \frac{1}{1 + \exp(i + 1 - \beta)}$ --- sharp decay (around breakout point $\beta$),
    \item $\sigma_{ii} = \frac{1}{i^{0.1}}$ --- slow decay.
\end{enumerate}

\begin{figure}[!ht]
    \centering
    \includegraphics[width=\textwidth]{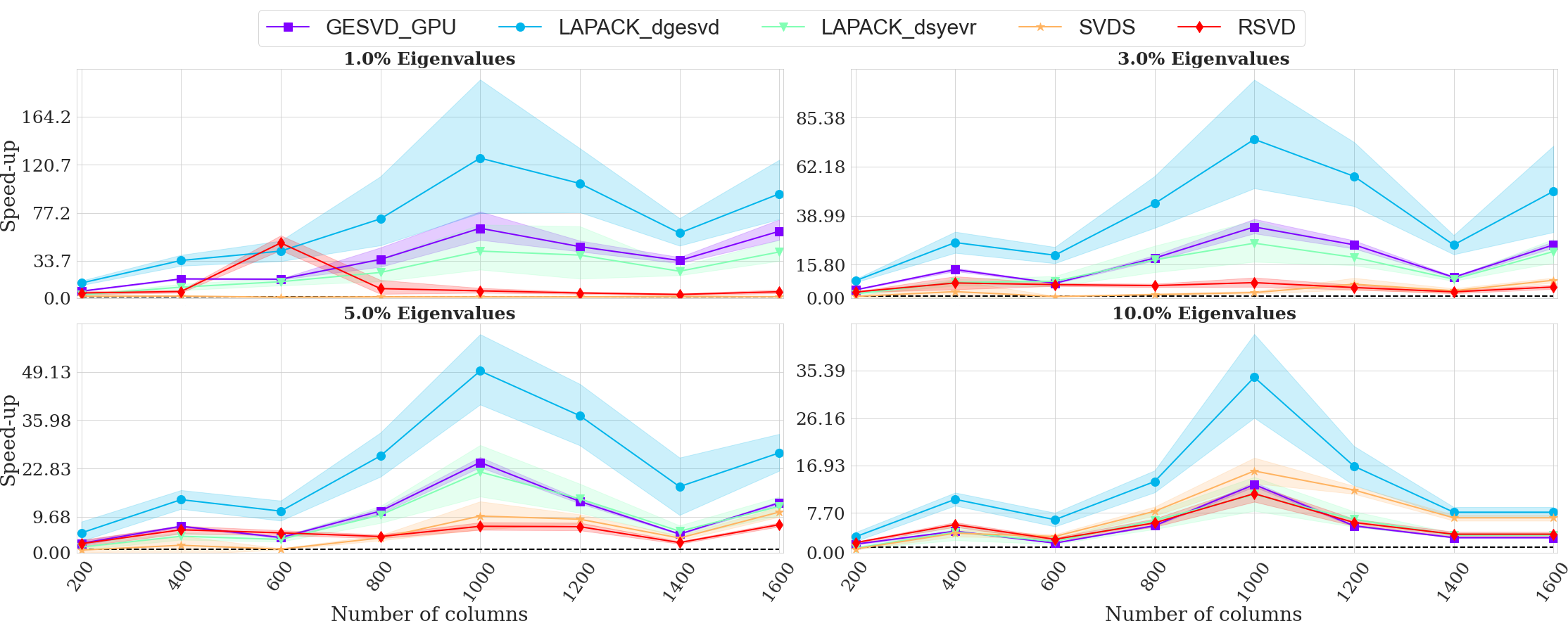}
    \caption{The speed-up of other methods to ours in the 'sharp decay' case of comparison. We show mean values of 10 runs as the line with standard deviation as the light areas. In this case we create matrix $A^{2000\times n}$, where $n$ is number of columns and we calculate 1\%, 3\%, 5\%, 10\% the largest eigenvalues for it.}
    \label{fig:perf_comp_decay_sharp}
\end{figure}

\begin{figure}[!ht]
    \centering
    \includegraphics[width=\textwidth]{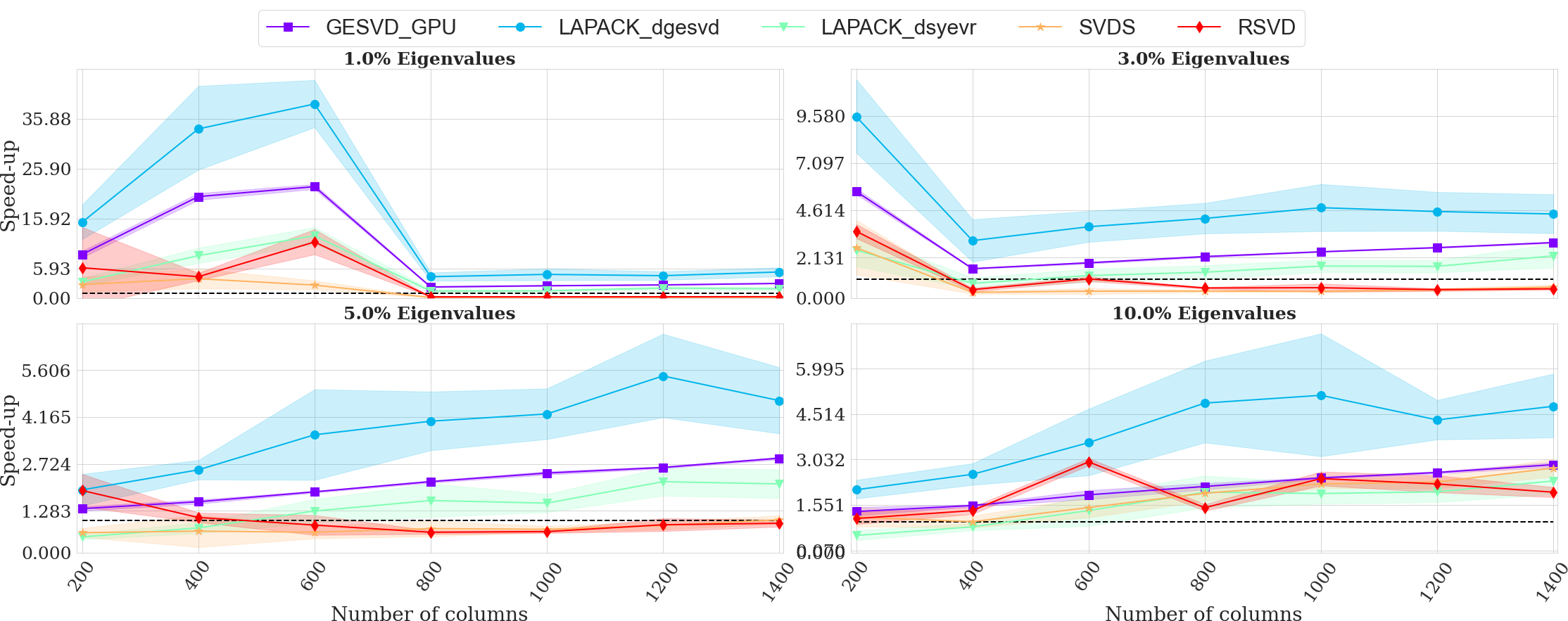}
    \caption{The speed-up of other methods to ours in the 'sharp decay' case of comparison. We show mean values of 10 runs as the line with standard deviation as the light areas. In this case we create matrix $A^{2000\times n}$, where $n$ is number of columns and we calculate 1\%, 3\%, 5\%, 10\% the largest eigenvalues for it. The dashed black line has a value of 1 (a reference to our method).}
    \label{fig:perf_comp_decay_slow}
\end{figure}

We take 1\%, 3\%, 5\%, 10\% of eigenvalues on matrix sizes i.e. for 5\% we calculate $k = \lceil 0.05n\rceil$ the largest eigenvalues. Due to the randomized singular value decomposition in our calculation, we kept the relative error on the limit of at most $10^{-8}$ against the baseline method, which is GESVD GPU. Figures~\ref{fig:perf_comp_decay_fast},~\ref{fig:perf_comp_decay_sharp},~\ref{fig:perf_comp_decay_slow} show the evaluation results. The dashed black line has a value of 1 (a reference to our method).

\paragraph{Application to Principal Component Analysis}

Principal Component Analysis (PCA) is a well-established technique for dimensionality reduction and multivariate analysis. Examples of its many applications include data compression, image processing, visualization, exploratory data analysis, pattern recognition, and time series prediction. 

PCA is related to eigenvectors and eigenvalues because with their help we can reduce the dimensionality of the data at a certain level of information preservation in the data. The eigenvectors of the covariance matrix of data are actually the directions of the axes where there is the most variance (most information) and that we call Principal Components. On the other hence, the eigenvalues are simply the coefficients attached to eigenvectors, which give the amount of variance carried in each Principal Component.

In this example, we compare a few state-of-the-art numerical methods to solve the eigenproblem. We consider not only the methods that can calculate the whole spectrum and associate with their eigenvalues. In particular, we focus on ones that can calculate only several eigenvalues, because our interest-only $k$ largest eigenvalues, i.e. we want to calculate only $1\%, 3\%, 5\%, 10\%, 20\%, 30\%$ Principal Components. Figure~\ref{fig:speedup_pca} shows the time of operation of other methods in relation to ours (speed-up). As a dataset, we take CelebA~\cite{liu2015faceattributes} where we resize the images to sizes $8\times8$, $12\times12$, $\ldots$, $52\times52$. Each RGB image about size $h\times w\times 3$, we flatten to the vector with size $3\cdot h\cdot w$, where $h$, $w$ is height, and the width of the image, respectively.

% \begin{figure}[!htb]
%     \centering
%     \includegraphics[width=\textwidth]{figures/speedup_pca_1.png}
%     \caption{Picture presents the time of operation of other methods in relation to ours (speedup). We consider two types of methods that calculate the whole spectrum and that calculate only $k$ largest eigenvalues.}
%     \label{fig:speedup_pca}
% \end{figure}

\paragraph{Application for subspace clustering}
Subspace clustering is one of the applications of the singular value decomposition of the matrices. In this part of the paper we present the results of comparison of the method SuMC that is presented in \cite{STRUSKI2018161} and application of this code with the randomized SVD algorithm that requires minor changes in the code.

\begin{table}[!ht]\small
\centering
\begin{tabular}{@{}ccccc@{}}
 {\bf Dataset} & {\bf Solver type} & {\bf Elapsed time (sec)} & {\bf Solver calls} & {\bf ARI score} \\
\toprule
\multirow{2}{*}{\em first} & CPU & 73653.0 & 262209 & 1.0 \\
& GPU & \phantom{0}2584.5 & 189291 & 1.0 \\
% \midrule
% \cmidrule(lr{.7em}){2-5}
% \hdashline[4pt/3pt]\noalign{\vskip 0.7ex}
\cdashline{2-5}[4pt/3pt]\noalign{\vskip 0.7ex}
\multirow{2}{*}{\em second} & CPU & - & - & - \\
 & GPU & 23967.0 & 1944144 & 1.0 \\
\bottomrule
\end{tabular}
\caption{Summary results of method SuMC which used solver of eigenvalues implemented on CPU and GPU in synthetic datasets.}
\label{tab:sumc}
\end{table}

Table~\ref{tab:sumc} shows the results of method of SuMC which used solver of eigenvalues implemented on CPU and GPU in synthetic datasets, which we randomly generate on the spaces $[0, 1]^{\text{dim}}$, where $\text{dim}$ denotes dimension of a space. 
In these generated data we known upfront what are the real subspaces and we can use the theoretical compression rate for the data set.
We generate 2 random synthetic datasets: the {\em first} contained $500$ points from $30$-dimensional subspace, $1000$ points from $50$-dimensional subspace and $2000$ points from $70$-dimensional subspace, where each points were presented in $1000$-dimensional space; the
{\em second} data contained $5000$ points from $30$-dimensional subspace, $10000$ points from $50$-dimensional subspace and $20000$ points from $70$-dimensional subspace (each points were described in $1000$-dimensional space), and for them, we calculated the mean of Rand index adjusted. In each experiment we started with the same initialization of points to clusters.

Note that the number of the solver call is the case of GPU in smaller than CPU which may indicate better generalization and this in turn result in faster convergence of the SuMC algorithm.

\section{Conclusion}
\label{sec:conclusions}

The experiments confirmed practical usability and high performance of the GPU implementation of a randomized SVD algorithm for computing small part of the spectrum. We have tested the method in three different tasks, i.e. just the performance analysis to compute 1-10\% of the spectrum, application for PCA, and application for subspace clustering. Obtained results were satisfactory across the board, with the speedup reaching up to 100x for some chosen cases. Be believe that applying this method for large scale PCA, as part of the machine learning pipeline might be instrumental.

\section*{Acknowledgments}
The authors would like to thank Lung Sheng Chien for discussion and comments on the paper. This research was supported by: grant no POIR.04.04.00-00-14DE/18-00 carried out within the Team-Net program of the Foundation for Polish Science co-financed by the European Union under the European Regional Development Fund. This research was funded in part by National Science Centre, Poland grant no.  2020/39/D/ST6/01332 and grant no. 2020/39/B/ST6/01511. For the purpose of Open Access, the authors have applied a CC-BY public copyright licence to any Author Accepted Manuscript (AAM) version arising from this submission.

% \section*{References}
\bibliographystyle{unsrt}
\bibliography{references}

\end{document}